\pdfoutput=1
\documentclass[review, 11pt]{article}
\makeatletter
\def\ps@pprintTitle{%
 \let\@oddhead\@empty
 \let\@evenhead\@empty
 \def\@oddfoot{\centerline{\thepage}}%
 \let\@evenfoot\@oddfoot}
\makeatother

\usepackage[margin=3cm]{geometry}

\usepackage{float}
\usepackage{lineno,hyperref}
\usepackage{amsmath}
\usepackage{amsfonts}
\usepackage{multirow}
\usepackage{textcomp}
\usepackage{graphicx}
\usepackage{booktabs}
\usepackage{authblk}
\usepackage{amssymb}
\usepackage{blindtext}
\usepackage{setspace}
\usepackage{array}
\usepackage{subcaption}

\DeclareSymbolFont{matha}{OML}{txmi}{m}{it}
\DeclareMathSymbol{\varv}{\mathord}{matha}{118}

\hypersetup{
    colorlinks=true,
    pdfstartview=FitV,
    linkcolor=blue,
    filecolor=blue,      
    urlcolor=blue,
    citecolor=blue
}
\modulolinenumbers[1]

\providecommand{\keywords}[1]
{
  \small	
  \textbf{\textit{Keywords---}} #1
}

\begin{document}

\title{Deep learning for objective estimation of Parkinsonian tremor severity}

\author[1]{Felipe Duque-Quiceno MSc} 
\author[1,*]{Grzegorz Sarapata MSc} 
\author[1]{Yuriy Dushin MSc} 
\author[1]{Miles Allen MSc}
\author[1,*]{Jonathan O'Keeffe MD, PhD} 

\affil[1]{Machine Medicine Technologies Ltd., The Biscuit Factory Unit J112, 100 Drummond Road, London SE16 4DG, UK}
\affil[*]{Corresponding authors: {greg, jonathan}@machinemedicine.com}
\date{September 03, 2024}
\maketitle
\doublespacing
\clearpage

\begin{abstract}
Accurate assessment of Parkinsonian tremor is vital for monitoring disease progression and evaluating treatment efficacy. We introduce a pixel-based deep learning model designed to analyse postural tremor in Parkinson's disease (PD) from video data, overcoming the limitations of traditional pose estimation techniques. Trained on 2,742 assessments from five specialised movement disorder centres across two continents, the model demonstrated robust concordance with clinical evaluations. It effectively predicted treatment effects for levodopa and deep brain stimulation (DBS), detected lateral asymmetry of symptoms, and differentiated between different tremor severities. 

Feature space analysis revealed a non-linear, structured distribution of tremor severity, with low-severity scores occupying a larger portion of the feature space. The model also effectively identified outlier videos, suggesting its potential for adaptive learning and quality control in clinical settings.

Our approach offers a scalable and objective method for tremor scoring, with potential integration into other MDS-UPDRS motor assessments, including bradykinesia and gait. The system's adaptability and performance underscore its promise for high-frequency, longitudinal monitoring of PD symptoms, complementing clinical expertise and enhancing decision-making in patient management. Future work will extend this pixel-based methodology to other cardinal symptoms of PD, aiming to develop a comprehensive, multi-symptom model for automated Parkinson's disease severity assessment.

\end{abstract}


\clearpage

\section*{Introduction}
\label{sec:introduction}
Tremor is a debilitating and core feature of Parkinson's disease and Essential Tremor, and can occur in numerous other neurological disorders \cite{louis2021prevalence,dorsey2018prevalence,willis2022incidence}. Its severity not only reflects disease progression but also serves as a biomarker for clinical decisions, such as determining the need for deep brain stimulation surgery or adjusting medication dosages \cite{friedrich2024validation}. Given its clinical significance, diagnosing and assessing tremor severity using highly objective, reliable and scalable measures is crucial.

In Parkinson’s disease, the Motor Disorder Society sponsored Revision of the Unified Parkinson’s Disease Rating Scale (MDS-UPDRS) \cite{UPDRS} is the standard tool for assessing disease severity, encompassing both motor and non-motor symptoms. Part III of this scale is dedicated to motor dysfunction, including an ordinal 5-point scale for tremor, among other symptoms. While this scale offers a structured approach to quantifying tremor severity based on amplitude, clinical ratings often suffer from inter-rater variability \cite{martinez1994unified,richards1994variability}. This variability stems from the inherent subjectivity in real-time clinical assessments, which depend heavily on the clinician's expertise and interpretation rather than on consistent, reproducible measures of tremor characteristics. Developing an automated solution that can provide comparable information through algorithmic analysis would offer clinicians a more reliable and objective tool, enhancing the assessment of disease progression and informing clinical decisions.

The widespread adoption of smartphones and tablets equipped with high-quality cameras has made the video recording of neurological motor assessments a common clinical practice. This trend has resulted in an ever-expanding dataset of multi-site, human-annotated assessments. Concurrently, the increasing popularity of artificial intelligence (AI), with its potential to learn complex patterns from data, has spurred the development of video-based models for motor disease screening. 

\begin{figure}[H]
    \centering
    \begin{tabular}[t]{c}
\begin{subfigure}{0.6\textwidth}
    \centering
    \smallskip
    \includegraphics[width=\linewidth]{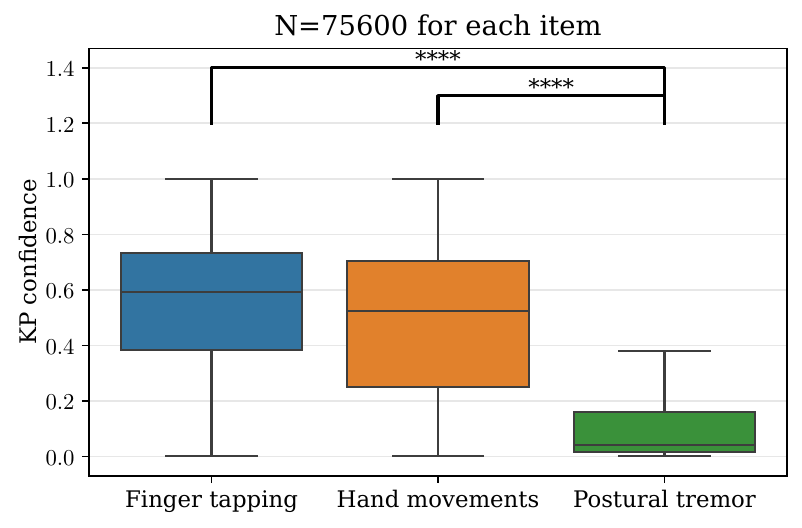}
    \caption{} 
    \label{subfig:kp_confidence}
\end{subfigure}\\
        \begin{tabular}[t]{c c}
        \smallskip
            \begin{subfigure}[t]{0.4\textwidth}
                \centering
                \includegraphics[width=0.9\textwidth]{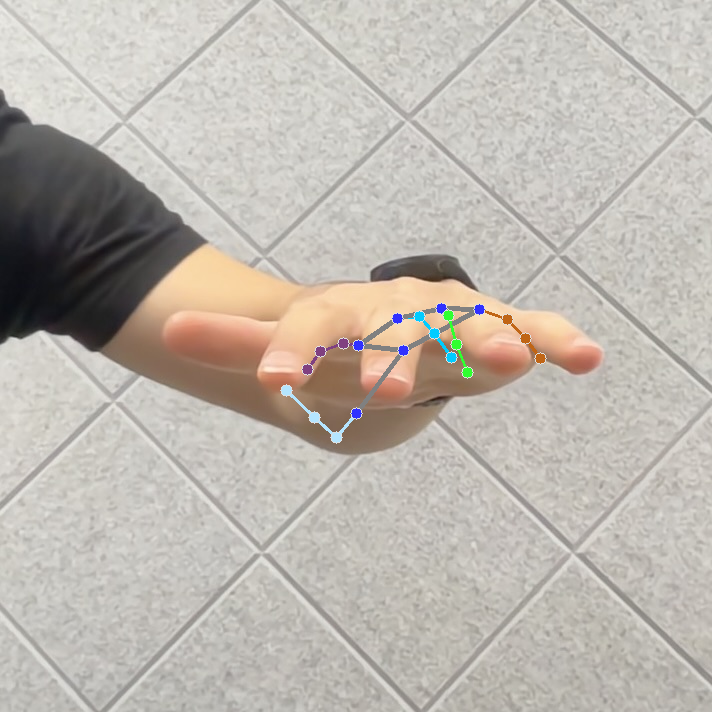}
                \caption{}
                \label{subfig:pth_position}
            \end{subfigure}
            \begin{subfigure}[t]{0.4\textwidth}
                \centering
                \includegraphics[width=0.9\textwidth]{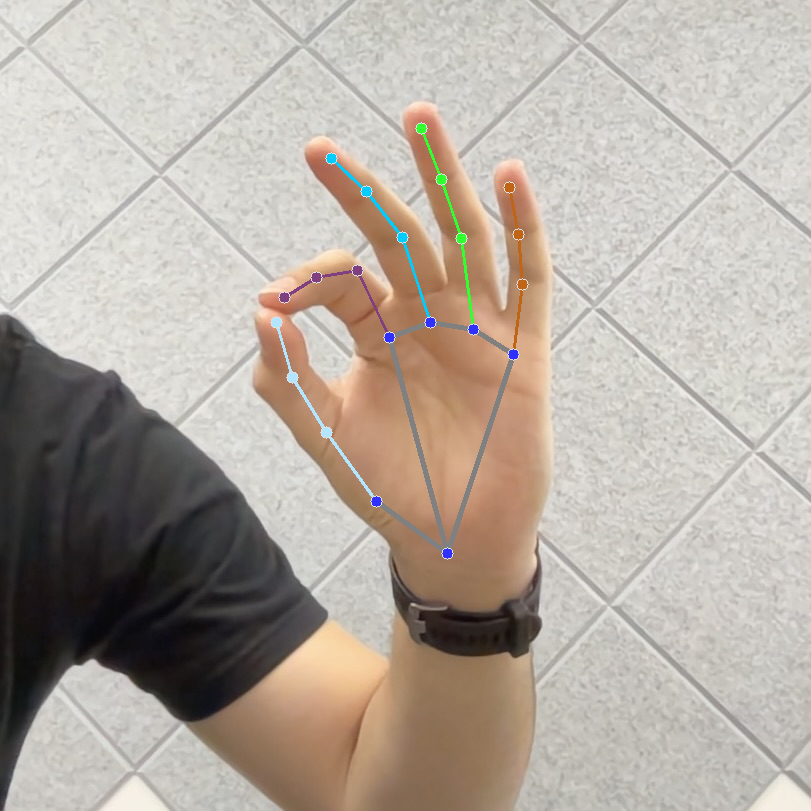}
                \caption{}
                                \label{subfig:ft_position}

            \end{subfigure}\end{tabular}\\
    \end{tabular}
    \caption{(a) Marker-less pose estimation confidence of the detected hand key points, for different MDS-UPDRS items. One-sided t-test comparison determined that postural tremor of hands had a significantly lower key point confidence ($\text{p-value}<0.0001$) than finger tapping and hand movements items. A total of 60 assessments were randomly sampled for each item, (12 from each severity score). Confidence from all hand key points ($n=21$) was extracted for a sub sample of 60 frames from each of the sampled assessments, accounting for a total of $n=75600$ confidence sample points for each item. (b)  Hand position for postural tremor assessment, as per MDS-UPDRS guideline \protect{\cite{UPDRS}}; occlusion of hand key points influences the confidence of the pose estimation model. (c) Hand key point visibility is less affected during other MDS-UPDRS items (finger tapping depicted), increasing the confidence of the predicted poses.
}
\label{fig:pose_estimation}
\end{figure}

In the context of PD, video-based algorithms have shown promise in severity scoring for other cardinal symptoms, utilising marker-less pose estimation methods \cite{morinan2023brady,morinan2022computer,rupprechter2021clinically,deng2024mediapipe}. Efforts have also been made to apply similar approaches to tremor analysis \cite{friedrich2024validation,Zhang2024,wang2021deep}. However, challenges such as video blur \cite{li2022blur} and occlusion of body landmarks (Figure \ref{fig:pose_estimation}) have limited the reliability of pose estimation methods for tremor scoring in clinical practice.

In this study, we aimed to develop a highly functional and robust model for parkinsonian tremor analysis that minimises reliance on precise body landmarks. Our approach leverages a deep learning model to extract kinematic patterns from spatio-temporal regions of interest in raw video data. Our results demonstrate the superior performance of this method compared to pose-dependent alternatives. Furthermore, we provide a thorough analysis of the model, illustrating its performance in clinically relevant scenarios and elucidating the features that contribute to its scoring decisions. Finally, we emphasise the scalability of our approach, highlighting its potential for widespread adoption in automatic motor dysfunction assessments.

\section*{Results}
\label{sec:results}
\subsection*{Comparison between models}
Across all the folds of a 5-fold cross-validation, the proposed 3D Conv-LSTM consistently outperformed the RFC model, measured on the test set (Table \ref{tab:model_comparison}), demonstrating the superiority of the pixel-based approach ($\text{p-value}=0.00312$) over the pose-reliant method. Notably, similar features and signals extracted for the RFC model have been successfully employed by the authors in training other, higher-performing models \cite{morinan2023brady}. This suggests that the RFC model's inferior performance is likely attributable to the video conditions rather than the model's implementation.

Out of the 2,742 assessments used, four involved corrupted videos and were thus considered training failures for both models. Additional failures were linked to instances where the pose estimation model failed to detect the necessary key points for an effective data processing. It is important to note that while the 3D Conv-LSTM model relied on pose estimation to extract hand bounding boxes, it only used central tendency measures from a few key points, making it less dependent on per-frame key point detection. In contrast, the RFC model required precise pose estimation in every frame to correctly extract signals and features, resulting in 417 more failed assessments compared to the pixel-based counterpart.

\subsection*{Severity prediction}
In specific contexts, such as the longitudinal monitoring of patients, binary classification of tremor severity can be advantageous. This allows for streamlined patient management by categorizing symptoms into distinct severity ranges. We assessed the model's performance across several binary classification thresholds, as follows: 
\begin{enumerate}
    \item \{0\} vs. \{1,2,3,4\}
    \item  \{0,1\} vs. \{2,3,4\}
    \item \{0,1,2\} vs. \{3,4\}\\
In practice, different thresholds can represent various clinical decision points, such as adjusting medication dosage or evaluating the need for DBS surgery.
\end{enumerate}

\begin{figure}[H]
 \begin{tabular}[t]{c c}
 \begin{subfigure}{0.55\textwidth}
\includegraphics[width=\linewidth]{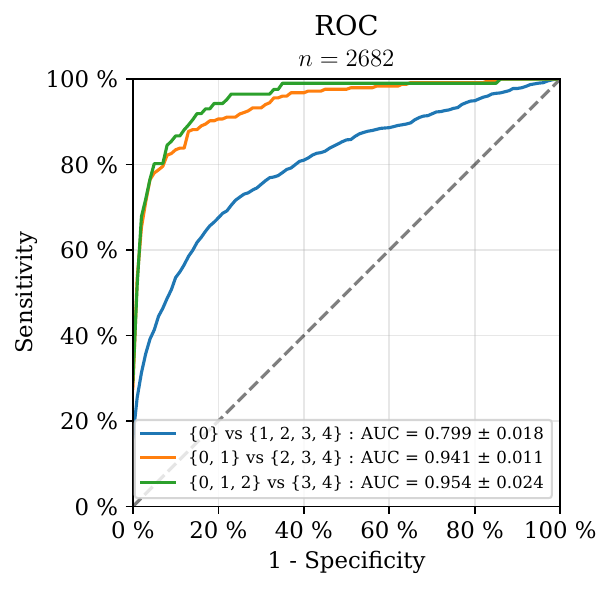}
\caption{}
\label{subfig:tremor_detection}
\end{subfigure}
&
\begin{subfigure}{0.5\textwidth}
\includegraphics[width=\linewidth]{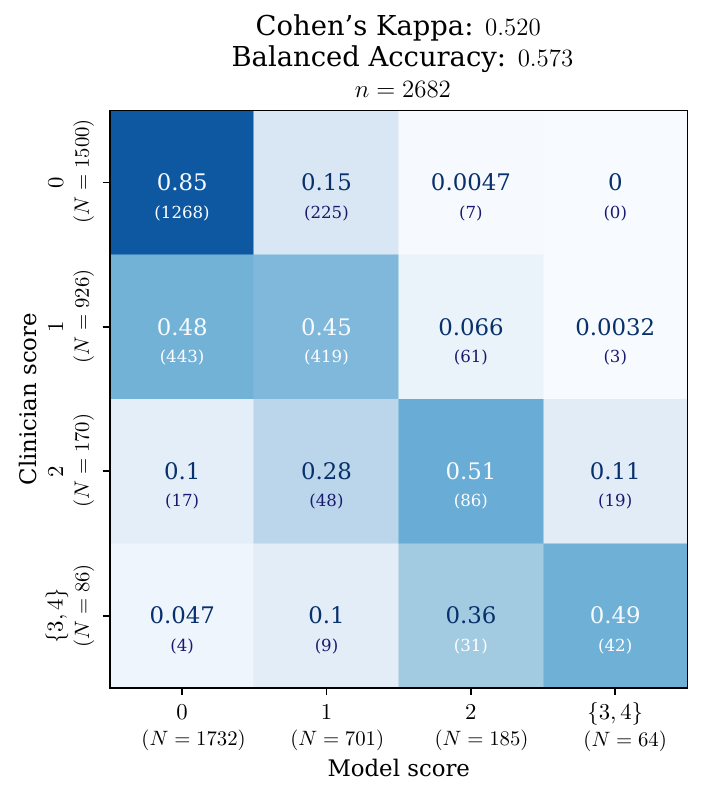}
\caption{}
\label{subfig:tremor_classification}
\end{subfigure}
 \end{tabular}
 \caption{(a) Performance of the 3D Conv-LSTM model across various binary classification tasks, measured by the area under the ROC curve (AUC). The gray dashed line indicates an AUC of 0.5, representing random behavior equivalent to flipping a fair coin. (b) Model performance in predicting the MDS-UPDRS-III item 3.15, measured using linearly weighted Cohen's Kappa and balanced accuracy on the combined test sets from a 5-fold cross-validation.}
\label{fig:binary_and_full_scoring}
\end{figure}

The area under the ROC curve (AUC) achieved in these tasks varied across thresholds (Figure \ref{subfig:tremor_detection}), with the first task achieving an AUC of 0.799, and a substantial improvement in the other two tasks (0.941 and 0.954, respectively). In practical terms, these results suggest that in a clinical scenario---such as detecting the critical point at which a patient’s tremor progresses from mild (e.g., score lower than 2) to intense---the proposed model can accurately identify this transition with a sensitivity of 76\%, when anchored to a specificity of 95\%.

The model achieved a linearly weighted Cohen's Kappa of 0.520 for full-range severity prediction (Figure \ref{subfig:tremor_classification}), where a value of 0.0 would indicate chance performance, equivalent to randomly sampling scores from the test set's distribution. A small proportion of assessments ($n=40$; approximately 1.5\%) were predicted outside of the $\pm$ 1 range from the clinical score. In practice, variability within this range is not uncommon among different raters \cite{goetz2010teaching}. Upon further examination, three out of the four most extreme misclassifications, with the clinical score of \{3, 4\} but the model prediction at 0, were associated with cases where the tremor occurred outside the temporal ROI accessible to the model.

\subsection*{Tremor asymmetry}
\begin{figure}[H]
\centering
\includegraphics[width=0.6\textwidth]{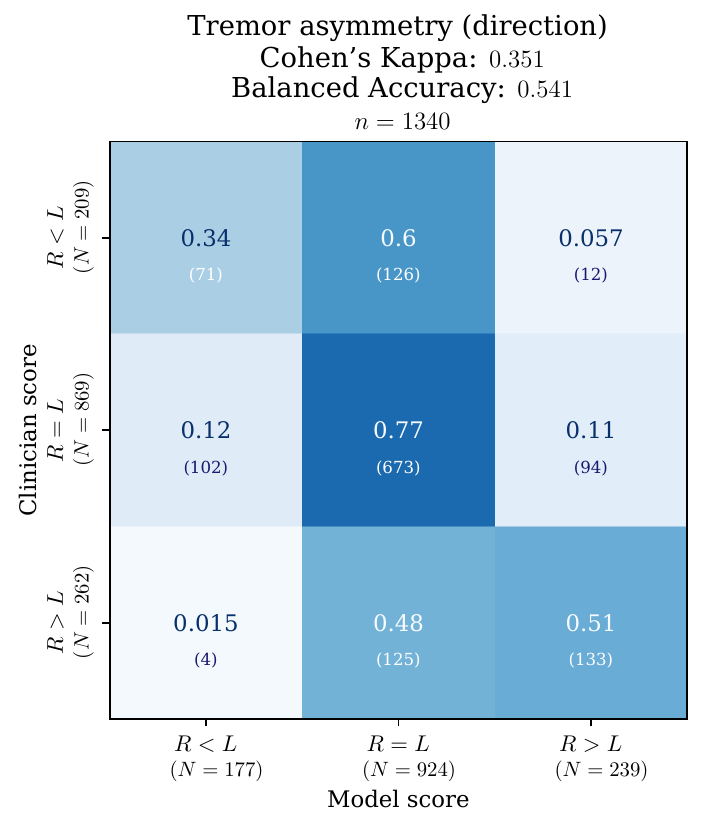}
\caption{Agreement between clinicians and the model to detect lateral asymmetry of tremor on $n=1340$ assessments with both left and right hand severity scoring. The model scores each hand independently, while clinicians have a access to both hands for a simultaneous comparison with one another. R: Right hand score, L: Left hand score}
\label{fig:asymmetry}
\end{figure}

For the patients with lateral asymmetry of symptoms ($n=471$, top and bottom rows in Figure \ref{fig:asymmetry}), the model incorrectly identified the direction of asymmetry in approximately 3.4\% of cases ($n=16$). Notably, 19\% of these misclassifications ($n=3$) occurred when tremor was present outside the temporal ROI as defined by the MDS-UPDRS guidelines. This model's ability to maintain the correct direction of asymmetry is noteworthy, especially given that it treats each laterality as an independent assessment, in contrast to clinicians who typically evaluate both hands simultaneously, allowing the severity in one hand to influence the scoring of the other.

Additionally, the model correctly identified symmetry in 77\% of assessments with lateral symmetry. However, it also misplaced a substantial proportion of asymmetric assessments into the symmetric category. Future work could investigate this behavior in greater detail to determine whether the model would benefit from bilateral training. Such an approach might enhance the model's ability to access mutual information between hands, potentially improving its capacity to more accurately define and detect asymmetry.

\subsection*{Effect of external stimulus}
\begin{figure}[H]
\centering\noindent\makebox[\textwidth]{\includegraphics{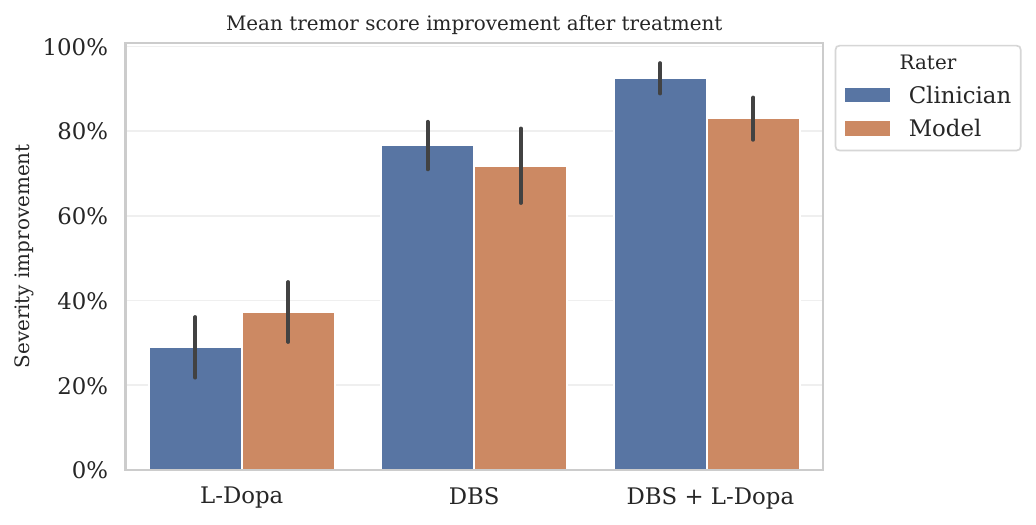}}
\caption{Average improvement in MDS-UPDRS-III postural tremor score after treatment relative to the baseline, $n=27$ patients with existent baseline tremor ($\mathrm{score} \geq  1$). Error bars indicate the standard error of the mean. Refer to tables \ref{tab:wilcoxon_between_treatments},\ref{tab:wilcoxon_between_raters} for statistical analysis of the different treatments. L-Dopa: levodopa, DBS: deep brain stimulation}
\label{fig:stimulation_effect}
\end{figure}

For a held-out cohort of 27 patients ($n=168$ assessments) with baseline tremor greater than zero, we evaluated the impact of various treatment modalities---levodopa, deep brain stimulation, and their combination---on predicted tremor severity (Figure \ref{fig:stimulation_effect}). Across all conditions, there was no significant difference between the model's predicted improvements and those observed by clinicians (Table \ref{tab:wilcoxon_between_raters}), suggesting that the model's sensitivity to treatment effect is on par with clinical assessments. Additionally, both the model and clinicians identified differences between therapies of similar significance (Table \ref{tab:wilcoxon_between_treatments}), including an approximately 70\% improvement with DBS, consistent with documented expectations \cite{krack1998dbs}.


\subsection*{Learned features}
\subsubsection*{t-SNE}
A two-dimensional embedding of the feature space was done to visually inspect the model's representation of tremor, using t-distributed stochastic neighbour embedding (t-SNE) \cite{vandermaaten2008tsne}. Considering that the principle behind this method is to maintain the relative proximity between samples from their high-dimensional feature space into the low-dimensional embedding, we would expect clustering and overall spacing between samples to be qualitatively informative. 

\begin{figure}[H]
\centering
\begin{tabular}{c}
\begin{subfigure}{\textwidth}
\centering
\includegraphics[width=0.8\textwidth]{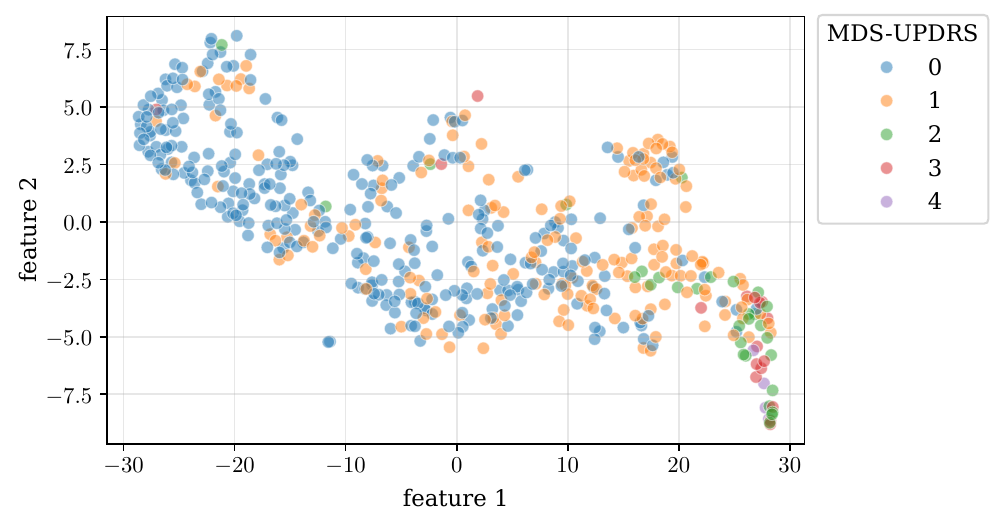}
\caption{}
\end{subfigure}\\
\begin{subfigure}{\textwidth}
\centering
    
\includegraphics[width=0.8\textwidth]{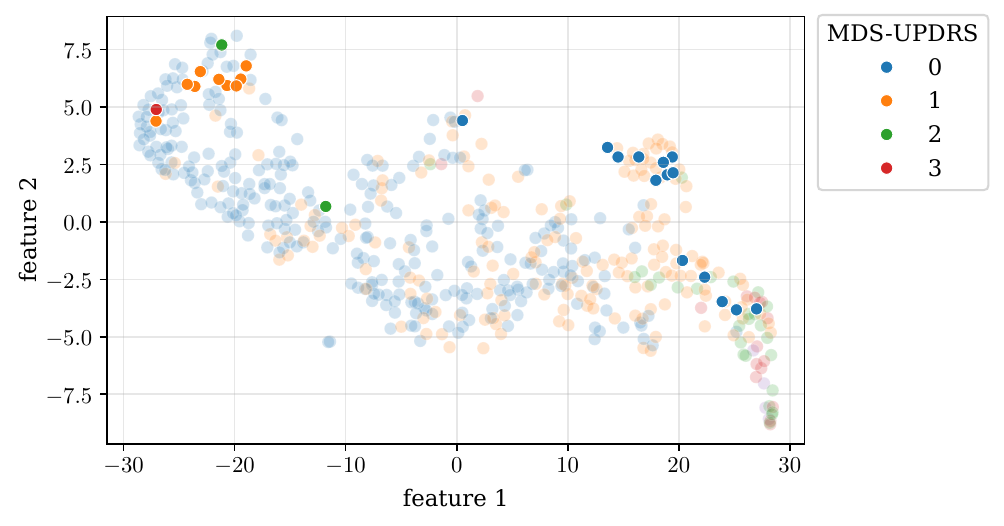} 
\caption{}
\end{subfigure} 
\end{tabular}

\caption{(a) Embedding of tremor severity in the model's feature space,  for the test dataset ($n=547$) of the first cross validated fold; t-SNE dimensionality reduction from 16 to 2 dimensions.(b) Highlight of the outliers for each class.}
\label{fig:tsne}
\end{figure}

Results revealed a clear structure in the model's representation of tremor, with increasing severity encoded along a gradient, where low severity scores clustered toward the upper left and high scores toward the lower right (Figure \ref{fig:tsne}). The proportion of feature space allocated to encoding low severity scores (0 and 1) was substantially larger than that for higher scores. This observation contrasts with the MDS-UPDRS scale, where the difference in tremor amplitude between scores of 0 and 1 is smaller than that between scores of 1 and 2, and so on, reflecting a non-uniform separation between severity levels \cite{UPDRS,elble2006logscale}. The observed disparity results logical when considering that the closer spacing between scores implies the need for more nuanced differentiation, explaining a larger portion of the feature space allocated to distinguishing these similar features. Conversely, higher severity scores, having a higher separation from one another, are easier for the model to classify, thus requiring less feature space. This is one possible explanation, but further analysis is required to reach a definitive conclusion.

A visual inspection of the outliers of each class's embedding, illuminated the model's logical structure in the distribution of severities across its feature space. Low severity samples ($\text{MDS-UPDRS}=0$) embedded in the high severity region ($n=13$) included: assessments with hand held camera recordings ($n=3$), which induced artificial movements to the detected image of the patient; malfunctioning camera auto-focus ($n=3$), creating a pulsating zoom-in and out throughout the video; non-tremor related hand movements ($n=1$) such as fidgeting of the fingers during the assessment. Similarly, high severity outliers ($\text{MDS-UPDRS}\geq2$, $n=3$) included two assessments where the tremor occurred outside of the MDS-UPDRS defined ROI, and one assessment where the tremor was limited to a single finger.

These results reveal the model's internal representation of tremor severity within a continuous abstract space, where movement intensities are distributed non-linearly, allowing low and high intensity movements to occupy distinct regions. This structured distribution not only reflects the model's ability to differentiate between varying severities but also highlights its potential for scalable and streamlined management of tremor patients. By identifying assessment videos that deviate significantly from their expected locations in the feature space, the model can flag these outliers for re-training purposes or, in a clinical setting, to identify potentially low-quality or mislabeled assessments.


\section*{Discussion}
\label{sec:discussion}
In this study, we successfully developed a highly functional and robust model for analysing parkinsonian tremor that reduces dependence on precise body landmarks. By employing a deep learning approach to extract kinematic patterns directly from raw video data, we demonstrated that our model outperforms traditional methods reliant on pose estimation. Our analysis underscores the model’s efficacy in clinically relevant scenarios and highlights its scalability for broad application in the streamlined assessment of motor dysfunction.

While tremor might appear easier to assess through video recordings compared to other cardinal symptoms of PD such as bradykinesia and rigidity, significant challenges remain. Rigidity requires direct physical interaction for accurate evaluation, and bradykinesia is difficult to assess due to its subjective interpretation, particularly when identifying subtle hesitations. In contrast, tremor severity is often more objectively measurable by assessing the amplitude and periodicity of movements. However, accurately estimating tremor using consumer-grade, non-specialised hardware remains challenging.

In the field of automatic parkinsonian tremor detection, substantial research has explored a wide variety of classification methods. Several studies have investigated the use of dedicated hardware to extract relevant signals \cite{jeon2017wearable,lemoyne2013wearable,alam2016inertial,pan2012lfp,vivar2019lmc}. While promising, these methods could impose additional costs on healthcare providers or patients and potentially complicate the already time-consuming MDS-UPDRS assessment process. Other approaches have examined the use of built-in inertial sensors in general-purpose smartphones \cite{fraiwan2016app,lemoyne2010app}; however, these methods require the patient to hold the device in their hands, which can interfere with the natural presentation of tremor symptoms.

Within the realm of computer vision, two main branches have emerged: pixel-based and markerless pose estimation approaches. Pixel-based methods, though less explored, include attempts such as the use of optical flow \cite{williams2021optical}, which tracks movement between frames by following corner points. While straightforward, this method has a high false positive rate due to its sensitivity to background noise \cite{wang2021deep}, rendering it unsuitable for real-world clinical settings. Conversely, markerless pose estimation approaches have shown greater success, with promising results in tremor detection reported by Friedrich et al. \cite{friedrich2024validation}, Zhang et al. \cite{Zhang2024}, and Wang et al. \cite{wang2021deep}. Nevertheless, these methods which rely on per-frame estimation of body landmarks face significant challenges---such as those previously discussed---when applied across diverse clinical environments. Moreover, they often depend on hand-crafted features that emphasise interpretability, which contrasts with the more data-driven approach advocated in this work.

Our work introduces an objective, reproducible, and scalable method for tremor scoring that is less vulnerable to errors and imprecision associated with pose estimation. Furthermore, our model's architecture is item-agnostic; although it was trained specifically for postural tremor in this study, it can be adapted to other PD-related motor assessments with minimal modifications.

Given the model's strong performance in clinically relevant scenarios---such as detecting treatment effects and identifying critical points in symptom progression---, our approach offers a promising solution for longitudinal and high-frequency monitoring of patients' conditions. This system is not intended to replace clinicians but to complement their expertise by providing a new stream of data, capable of capturing daily severity cycles and monitoring disease progression at home. Clinicians could leverage this information to enhance decision-making and gain deeper insights into the complexities of motor impairments in Parkinson’s disease.

Moreover, by exploiting the item-independence of this approach, tremor kinematics can be integrated with those of bradykinesia, gait and arising from chair, if developed as a part of the same system. This study is confined to the detection and analysis of tremor, a significant but singular cardinal symptom of PD, but we anticipate that employing pixel-based methodologies, as described here, could similarly benefit models designed for other MDS-UPDRS items. Future research should explore extending the pixel-based approach to encompass a broader range of MDS-UPDRS components. By adopting this approach across multiple symptoms, it is anticipated that models can achieve enhanced accuracy and broader applicability in assessing PD severity.

The ultimate objective is to develop a foundational model capable of predicting severity across multiple MDS-UPDRS items, rather than limiting the focus to individual subsets. This broader model would aim to encapsulate a comprehensive embedding of Parkinsonism kinematics, allowing for fine-tuning with potentially smaller datasets. This approach mirrors the adaptability seen in general-use markerless pose estimation models, which can be efficiently tuned to extract poses across various scenarios. By extending this capability to cover a wide range of MDS-UPDRS assessments, the envisioned model could significantly enhance automated assessment techniques for Parkinson's disease. This strategic direction seeks to advance the field by establishing a unified framework for automated Parkinson's disease severity assessment, leveraging learned kinematic embedding to improve predictive capabilities across a broad spectrum of clinical symptoms.
\clearpage

\section*{Methods}
\label{sec:methods}

\subsection*{Subjects and assessments}
MDS-UPDRS postural tremor of hands assessments were conducted for PD patients at five specialised movement disorders centres located in the United Kingdom and the United States, resulting in a total of 2,742 assessments---1,371 of each laterality (Tables \ref{tab:demographics},\ref{tab:updrs_distribution}). The severity of postural tremor was assessed using a standardised 5-point scale (0 to 4) as specified in the MDS-UPDRS Part III protocol. Video recordings and evaluations were performed using KELVIN™, a video-based motor assessment platform developed by Machine Medicine Technologies \cite{MMT, sibley2021video}. This platform has been utilised in previous studies involving various aspects of the MDS-UPDRS \cite{updrs8mmt, morinan2022computer}. 

The videos captured a broad spectrum of disease severities, including both ON and OFF states for medication and deep brain stimulation conditions, and were recorded using consumer-grade cameras integrated into mobile devices or tablets. Approximately 90\% of the recordings were conducted at a resolution of 1080x1920 pixels and a frame rate of 29.97 frames per second, typical settings found on modern mobile devices.

Videos underwent automatic filtering based on criteria such as minimum length and frame rate. No manual selection of videos occurred, ensuring that the dataset accurately represents routinely collected clinical data at these sites. Clinicians were only advised to use a tripod and ensure the patient remained fully visible and centred within the video frame throughout the assessment. Only a single frontal view was recorded for each assessment. This study specifically focused on item 3.15 of the MDS-UPDRS, which evaluate the severity of postural tremor in the hands. 

\subsection*{Training pipeline}
The inference data pipeline is composed of two primary components: video pre-processing and the machine learning model. During pre-processing, the video was cropped to a region of interest (ROI) both spatially and temporally. Although previous work by this group has demonstrated the capability to automatically detect temporal ROIs for various PD-related activities \cite{sarapata2023activity},  in this study, the temporal ROI was manually labelled to avoid error propagation between models. The start and finish frames of the ROI were defined as the first and last frame where the patient hand their arms stretched out in front of the body with the palms down, as per MDS-UPDRS guidelines \cite{UPDRS}; other non-official positions were excluded from the region of interest.

The spatial ROI was defined as a square bounding box centred at the median position of the wrist across the temporal ROI, with the side length proportional to the patient's mean spine length (refer to Table \ref{tab:preprocessing} for details). The resulting hand bounding box was then resized to a 32x32 pixel colour image, to reduce the dataset size, thereby optimising computational efficiency.

Marker-less pose estimation was used exclusively for defining the spatial ROI, with its dependence on precise frame-by-frame estimation minimised by using only central tendency measures of the required key points. Mediapipe landmark detection models \cite{deng2024mediapipe} were employed to extract these key points.

Subsequently, the processed hand bounding box was fed into the machine learning component. We implemented a 3D Convolutional Network followed by a bidirectional Long Short-Term Memory (LSTM) network with an attention layer, forming the 3D Conv-LSTM architecture. The convolutional layers were intended to learn the embedding of local spatio-temporal features characteristic of PD kinematics, while the LSTM layers were intended to classify these features into corresponding severity levels based on their temporal dynamics (See Table \ref{tab:model_params} for a detailed list of the model's layers and parameters).

\subsubsection*{Training and validation}
The training data was divided into five splits, stratified by MDS-UPDRS scores to ensure a similar distribution of labels across splits, and grouped by assessment ID to keep information from the same patient within the same split. A 5-fold cross-validation was then performed, with three splits used for training, one for validation, and one for testing. Due to the dataset's significant imbalance toward lower severity scores (as shown in Table \ref{tab:updrs_distribution}), severities 3 and 4 were merged into a single category for training purposes. This approach effectively trained the model to classify assessments into one of four categories: \{0, 1, 2, \{3, 4\}\}. Additionally, to address the imbalance, minority classes were over-sampled during model training.

The model was trained for a total of 250 epochs using a reduce-on-plateau scheduler, with a patience of 15 epochs and a reduction factor of 10. Early stopping was triggered when the learning rate reached $5\times10^{-6}$. The Adam optimiser \cite{KingBa15} was employed, starting with an initial learning rate of 0.0006 and a weight decay of 0.001. Additionally, data augmentations, including colour and affine transformations, were applied to the training set (Table \ref{tab:augmentations}).

For comparison against pose-reliant methods, a Random Forest Classifier (RFC) was trained using hand-crafted pose features, with the same dataset and cross-validation folds. Refer to Table \ref{tab:rfc_features} for a detailed list of features. 

\subsection*{Statistical analysis}
\subsubsection*{Models comparison}
To compare the performance of the 3D Conv-LSTM and the RFC models, we tested the null hypothesis that both models perform equally ($H_0: \pi \leq 0.5$, where $\pi$ is the probability of the 3D Conv-LSTM outperforming the RFC in any given trial). A one-sided binomial test was conducted at a 5\% significance level on the test splits of the cross-validation, testing the alternative hypothesis that the pixel-based 3D Conv-LSTM model was superior to the pose-reliant RFC model, measured by the linearly weighted Cohen's Kappa.

\subsubsection*{Tremor score}
The primary evaluation metric was the linearly weighted Cohen's Kappa, to measure agreement between the clinician and model estimate of tremor score. Balanced accuracy (average recall obtained for each class) was chosen as secondary evaluation metric to address the dataset's class imbalance.

Linearly weighted Cohen's Kappa was also applied to evaluate the model’s performance in detecting tremor asymmetry, specifically whether the tremor severity in the right hand was lower, equal to, or higher than that in the left hand.

Additionally, the model was tested on binary classification of tremor at various threshold points (classifications 0 vs \{1,2,3,4\}, \{0,1\} vs \{2,3,4\} and \{0,1,2\} vs \{3,4\}). Model performance in this binary tasks was evaluated using the area under the receiver operating characteristic curve, which assesses the model's sensitivity and specificity in detecting tremor intensity. 

\subsubsection*{Effect of stimulation}
An additional held-out dataset comprising 168 assessments from 27 post-surgical DBS patients was collected from a clinical centre not involved in the training phase. All patients participated in a levodopa challenge test \cite{saranza2020levodopa} in both ON and OFF states of DBS stimulation, resulting in a total of three treatment combinations plus the baseline condition. The relative improvement in tremor severity for each type of treatment (levodopa, DBS, DBS + levodopa) was evaluated pairwise using a one-sided Wilcoxon signed-rank test at a 5\% significance level. Refer to Tables \ref{tab:wilcoxon_between_treatments},\ref{tab:wilcoxon_between_raters} for details. A baseline tremor score greater than 0 was established as inclusion criterion for this analysis.

\subsection*{Feature embedding}
To visualise the learned embedding of kinematic features, we applied t-SNE dimensionality reduction \cite{vandermaaten2008tsne} to the final layer of the LSTM, just before the fully connected prediction layer. This technique reduced the 16-dimensional feature space into a 2-dimensional plane. We embedded the test samples using a perplexity value of 50 and an early exaggeration factor of 30. An elliptic envelope was fitted to each class and their outliers were then visually inspected to better understand the potential limitations of the model's learned embedding and the factors contributing to uncertainty in its predictions.

\subsection*{Ethics statement}
Written informed consent was obtained from all subjects and the agreements formed with institutions providing the data to be used in this research.
Data was collected at: Department of Clinical and Movement Neuroscience, Institute of Neurology, University College London; 
Neuroscience Research Centre, Molecular and Clinical Sciences Research Institute, St. George’s, University of London;
Dementia Research Center, Institute of Neurology, University College London;
Parkinson’s Disease and Movement Disorders Center, Baylor College of Medicine; The Starr Lab, University of California San Francisco; Mater Misericordiae University Hospital, Dublin. Data was collected according to approval and ethics procedures of each participating institution.

Explicit written consent was obtained for publishing all patient photographs presented in this research.

\clearpage

\section*{Data availability}
The full data that support the findings of this study are not available for reasons of patient confidentiality and privacy. Anonymised summary data that support the findings of this study are available from the corresponding authors upon reasonable request.

\section*{Code availability}
Classification models were constructed using versions 2.3.0 of pytorch (\url{pytorch.org}) and 1.2.2 of scikit-learn (\url{scikit-learn.org}). 
Data processing, feature creation and statistical analysis used; version 2.1.1 of pandas (\url{pandas.pydata.org}), version 1.26.4 of numpy (\url{numpy.org}) and version 1.11.3 of scipy (\url{scipy.org}).
Data visualisations were created using version 3.9.0 of matplotlib (\url{matplotlib.org}) and  version 0.13.2 of seaborn (\url{seaborn.pydata.org}).

\section*{Acknowledgements}
This study was funded by Innovate UK and Machine Medicine Technologies. We thank the staff of research centers involved in the data collection. We also thank the employees at Machine Medicine Technologies for labelling the temporal regions of interest in all videos.

\section*{Competing interests}
The authors declare no Competing Non-Financial Interests but the following Competing Financial Interests: All authors contributed to this work as employees of Machine Medicine Technologies, owner of the Kelvin platform used in this research.

\section*{Author contributions}
Felipe Duque-Quiceno contributed to: research project organisation and execution, statistical analysis design and execution, manuscript first draft and review and critique.
Grzegorz Sarapata, Yuriy Dushin contributed to: research project organisation and execution, statistical analysis design and execution, manuscript review and critique.
Miles Allen contributed to: design and implementation of the RFC model, signals and feature extraction.
Jonathan O'Keeffe contributed to: research project conception and  organisation, statistical analysis design, manuscript review and critique.

\newpage

\footnotesize
\bibliography{mybibfile}
\newpage
\section*{Figures}

\newpage
\begin{figure}[H]
\centering
\includegraphics[width=\textwidth]{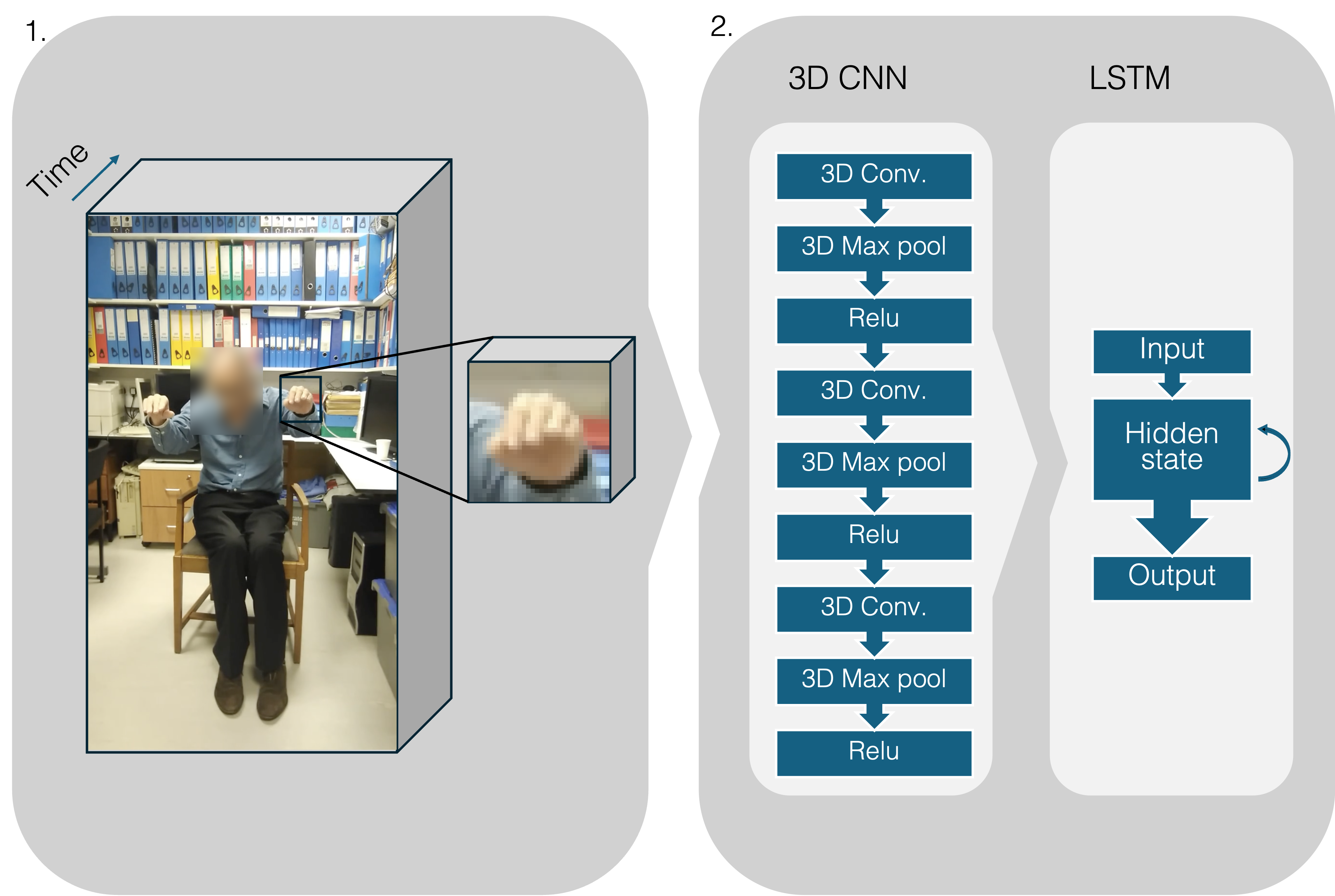}
\caption{Schematic of the inference pipeline. 1. Data pre-processing: Crop and resize of videos. 2. Inference model: Data is fed to a 3D CNN and then an LSTM module, finalizing with a fully-connected prediction layer. Refer to Table \ref{tab:model_params} model parameter details.}
\label{fig:pipeline}
\end{figure}
\newpage

\newpage

\newpage
\newpage

\newpage
\newcommand{\colWidth}{0.15\textwidth}

\begin{table}[H]
\footnotesize
    \centering
\begin{tabular}{cc|cc|c|c|c}
\multicolumn{2}{m{\colWidth}|}{\textbf{Age in years}}&\multicolumn{2}{m{\colWidth}|}{\textbf{Disease duration in years}}&\multirow{2}{\colWidth}{\textbf{Females proportion}}&\multirow{2}{\colWidth}{\textbf{DBS present}}&\multirow{2}{\colWidth}{\textbf{With medication}}\\
Mean&(SEM)&Mean&(SEM)&&&\\
\hline
61&(0.3)&8&(0.2)&31.6\%&30.0\%&56.5\%\\

    \end{tabular}
    \caption{Statistics summarising patient characteristics. Demographic data available for 892 out of the 2,742 total assessments, and assumed to follow the same distribution. SEM: standard error of the mean.}
    \label{tab:demographics}
\end{table}
\newpage

\begin{table}[H]
\footnotesize
    \centering
\begin{tabular}{c|c|c}
\textbf{MDS-UPDRS}&\textbf{n}&\textbf{\%}\\
\hline
0&1545&56.34\\
\hline
1&933&34.03\\
\hline
2&175&6.38\\
\hline
3&77&2.81\\
\hline
4&12&0.44\\
\hline\hline
\textbf{Total}&\textbf{2742}&\textbf{100\%}

    \end{tabular}
    \caption{Distribution of postural tremor of hands severity scores in the training dataset.}
    \label{tab:updrs_distribution}
\end{table}
\newpage

\begin{table}[H]
\footnotesize
    \centering
\begin{tabular}{c|cc|cc|cc|cc|cc}
& \multicolumn{2}{c|}{\textbf{Fold 0 (n=547)}}& \multicolumn{2}{c|}{\textbf{Fold 1 (n=548)}}&  \multicolumn{2}{c|}{\textbf{Fold 2 (n=548)}}&   \multicolumn{2}{c|}{\textbf{Fold 3 (n=552)}}&  \multicolumn{2}{c}{\textbf{Fold 4 (n=547)}}\\
         & K&n'& K&n'& K&n'& K&n'& K&n'\\
         \hline
         \multirow{2}{40pt}{3D Conv-LSTM} &  0.520& 532& 0.544&532&  0.497& 537& 0.524&543& 0.518&538\\
         &&&&&&&&&\\
         \hline
         RFC&  0.260&457&  0.374&453&  0.368&437&  0.380&461& 0.370&457\\
    \end{tabular}
    \caption{Test set performance comparison between the pixel based model 3D Conv-LSTM model versus the pose estimation-dependent RFC model. K: linearly weighted Cohen's Kappa; n': number of non-failing assessments. Failures occur when relevant key points could not be consistently detected throughout the video.}
    \label{tab:model_comparison}
\end{table}
\newpage
\newcommand{\stepRowWidth}{50pt}

\begin{table}[H]
\centering
\begin{tabular}{c|m{0.45\textwidth}|l}
{\textbf{Step}}&{\textbf{Definitions}}  & {\textbf{Formula}}\\
\hline
\multirow{9}{\stepRowWidth}{Mean spine-length detection}&ROI: Temporal region of interest  & \\
  &$\mathbf{p}_{\mathrm{sh}_i}$: Position vector of the ---left or right--- shoulder& \\
  &$\mathbf{p}_{\mathrm{hip}_i}$: Position vector of the ---left or right--- hip& \\
  &mean$(\cdot)$: Spatial average&\\
  &$\mathbf{p_\mathrm{neck}}$: Position vector of the neck& $\mathbf{p_\mathrm{neck}}=\mathrm{mean}(\mathbf{p}_{\mathrm{sh}_L},\mathbf{p}_{\mathrm{sh}_R})$\\
  &$\mathbf{p_\mathrm{m{\text -}hip}}$: Position vector of the mid-hip&$\mathbf{p_\mathrm{m{\text -}hip}}=\mathrm{mean}(\mathbf{p}_{\mathrm{hip}_L},\mathbf{p}_{\mathrm{hip}_R})$ \\
  & $\lVert \cdot\rVert$: Euclidean distance&\\
  &size$(\cdot)$: Number of elements in a set&\\
  &$h$: Mean spine length&$h=\frac{1}{\mathrm{size(ROI)}}\sum_{t\in \mathrm{ROI}}{\lVert\mathbf{p_\mathrm{neck}}-\mathbf{p_\mathrm{m{\text -}hip}}\rVert_t}$\\
  \hline
  \multirow{4}{\stepRowWidth}{Hand bounding box}&${\mathbf{p}}_\mathrm{wrist}$: Position vector of the wrist& \\
  &median$(\cdot)$: Median across time&\\
  &$b_{c}$: Center point of the bounding box&$b_{c}=\mathrm{median}({\mathbf{p}}_\mathrm{wrist})$\\
  &$b_{l}$: Side length of the bounding box&$b_{l}=0.593^\mathrm{\dag}h$\\
    \end{tabular}
    \caption{Pre-processing of the video data prior to being passed throught the deep learning model. \dag: value defined empirically, to maximize the coverage of the hand while minimizing the total size of the bounding box.}
    \label{tab:preprocessing}
\end{table}
\newpage
    

\begin{table}[H]
\centering
\begin{tabular}{c|c|c|cccccccc} 
    \multirow{2}{*}{\textbf{Section}}&\multirow{2}{*}{\textbf{Layer}}&\multirow{2}{*}{\textbf{Module}}&\multicolumn{8}{c}{\textbf{Parameters}}\\
    &&&$c_{\text{in}}$&$c_{\text{out}}$&$k_s$&$k_t$&$s_s$&$s_t$&$p_s$&$p_t$\\
    \hline
    \multirow{9}{*}{3D CNN}&1&Conv3d&3&32&3&5&1&1&0&2\\ 
    &2&MaxPool3d&32&32&2&1&2&1&0&0\\
    &3&Relu&&&&&&&\\ 
    &4&Conv3d&32&32&3&5&1&1&0&2\\ 
    &5&MaxPool3d&32&32&2&2&2&2&0&0\\
    &4&Relu&&&&&&&&\\ 
    &7&Conv3d&32&32&3&5&1&1&0&2\\  
    &8&MaxPool3d&32&32&2&2&2&2&0&0\\
    &9&Relu&&&&&&&\\ 
    \hline\hline
    \multicolumn{3}{c|}{}&$c_{\text{in}}$&$b$&$h_w$&$h_d$&$d$\\
    \hline
    LSTM&10-13&LSTM&32&True&8&3&0.1\\
    \hline\hline
    \multicolumn{3}{c|}{}&$h_w$&$c_{\text{out}}$\\
    \hline
    Prediction&14&Linear&16&4\\
    
    \end{tabular}
    \caption{List of parameters for the 3D Conv-LSTM model. $c_{\text{in}}$: input channels; $c_{\text{out}}$: output channels; $k_s$: spatial kernel size; $k_t$: temporal kernel size; $s_s$: spatial stride; $s_t$: temporal stride; $p_s$: spatial padding; $p_t$: temporal padding; $b$: bidirectional; $h_w$: number of features in the hidden state; $h_d$: number of hidden layers; $d$: dropout.}
    \label{tab:model_params}
\end{table}
\newpage
\newcommand{\firstRowWidth}{0.15\textwidth}
\newcommand{\secondRowWidth}{0.4\textwidth}

\begin{table}[H]
    \centering
    \small
    \begin{tabular}{m{\firstRowWidth}|m{\secondRowWidth}|l}
         \textbf{Name}&\textbf{Description}  &\textbf{Parameters} \\
         \hline
         Horizontal flip& Flips the video horizontally (\textit{Left} becomes \textit{Right}) with a probability of occurrence $p$.& $p=0.5$\\
         \hline
         \multirow{4}{\firstRowWidth}{Colour jitter}&\multirow{4}{\secondRowWidth}{Randomly change the brightness, contrast, saturation and hue of the video within the specified ranges.}&brightness: $[-10,+10]$ \%\\
         &&contrast: $[-10,+10]$ \%\\
         &&saturation: $[-30,+30]$ \%\\
         &&hue: $[-10, +10]$ \%\\
         \hline
         \multirow{4}{\firstRowWidth}{Affine transformation}&\multirow{3}{\secondRowWidth}{Performs a random rotation, translation and scaling to the video within the specified ranges.}&rotation: $[-30, +30]$ \textdegree\\
         &&translation: $[-10,+10]$ \% from centre\\
         &&scale: $[0.9,1.5]$
    \end{tabular}
    \caption{Augmentations used during training.}
    \label{tab:augmentations}
\end{table}
\newpage

\begin{table}[H]
\centering
\begin{tabular}{c|l|m{0.5\textwidth}}
\textbf{Group}&{\textbf{Name}}&{\textbf{Description}}\\
\hline
\multirow{2}{*}{Signals}&\textit{Vertical amplitude}&Distance between the peaks and troughs of the vertical component of the key point's position vector.\\
\cline{2-3}
&\textit{Frequency representation}&Discrete Fourier transform of the time-series position vector of a key point\\
\hline
\multirow{5}{*}{Features}&\textit{Mean amplitude}&Average \textit{vertical amplitude} of the key point movement across time\\
\cline{2-3}
&\textit{Max amplitude}&The 90th percentile of the key point's \textit{vertical amplitude}\\
\cline{2-3}
&\textit{Peak frequency}&Dominant frequency of the key point's \textit{frequency representation}\\
\cline{2-3}
&\textit{Mean frequency}&First moment of the key point's \textit{frequency representation}\\
\cline{2-3}
&\textit{Relative power of tremor}&Power of the signal in the range $\pm$6 Hz from the mean frequency, proportional to the total \textit{frequency representation} power\\

    \end{tabular}
    \caption{List of hand crafted signals and features used for training a random forest classifier. Signals were extracted from the raw key point positions; features were extracted from signals. Each feature was applied to the following key points: thumb, index finger, pinky finger, wrist, elbow and shoulder.}
    \label{tab:rfc_features}
\end{table}
\newpage

\begin{table}
    \centering \small
\begin{tabular}{c|c|c|c|c|m{0.1\textwidth}|c}
\multirow{2}{*}{\textbf{Rater}}&\multirow{2}{*}{\textbf{Treatment $x$}}&\multirow{2}{*}{\textbf{Treatment $y$}}&\multicolumn{4}{c}{\textbf{One-sided Wilcoxon signed rank test}} \\

 &  &  & Statistic & p-value & Corrected p-value& Significance\\
 \hline
         Clinician&L-Dopa&DBS&  18.0&  0.000038&  0.000338& ***\\
         \hline
         Clinician&  L-Dopa&  DBS + L-Dopa&  21.0&  0.000002&  0.000014& ***\\
         \hline
         Clinician&  DBS&  DBS + L-Dopa&  10.5&  0.006549&  0.058943& ns\\
         \hline
         Model&  L-Dopa&  DBS&  50.0&  0.000519&  0.004674& **\\
         \hline
         Model&  L-Dopa&  DBS + L-Dopa&  21.5&  0.000017&  0.000149& ***\\
         \hline
         Model&  DBS&  DBS + L-Dopa&  9.0&  0.054093&  0.486834& ns\\
        
\end{tabular}
\caption{One-sided Wilcoxon signed rank test, testing the alternative hypothesis that treatment $y$ shows greater improvement than treatment $x$. Used bonferroni correction for the $n=9$ comparison (these and table \ref{tab:wilcoxon_between_raters}). L-Dopa: levodopa; DBS: deep brain stimulation; *: $\text{p-value}<0.05$; **: $\text{p-value}<0.01$;***: $\text{p-value}<0.001$; ns: not significant.}
    \label{tab:wilcoxon_between_treatments}
\end{table}

\begin{table}
    \centering \small
\begin{tabular}{c|c|c|c|c}
\multirow{2}{*}{\textbf{Treatment}}&\multicolumn{4}{c}{\textbf{Two-sided Wilcoxon signed rank test}} \\

   & Statistic & p-value & Corrected p-value& Significance\\
 \hline
         L-Dopa&  80.0&  0.348379&  3.135411& ns\\
         \hline
         DBS&  0.0&  0.026857&  0.24171& ns\\
         \hline
         DBS + L-Dopa&  50.5&  0.899293&  8.093638& ns\\

\end{tabular}
\caption{Two-sided Wilcoxon signed rank test, testing the alternative hypothesis that a treatment is perceived to produce a different improvement by the clinician and the model. Used bonferroni correction for the $n=9$ comparison (these and table \ref{tab:wilcoxon_between_treatments}). L-Dopa: levodopa; DBS: deep brain stimulation; *: $\text{p-value}<0.05$; **: $\text{p-value}<0.01$;***: $\text{p-value}<0.001$; ns: not significant.}
    \label{tab:wilcoxon_between_raters}
\end{table}

\end{document}